\documentclass[runningheads]{llncs}
\usepackage{tikz}
\usepackage{multirow}
\usepackage{comment}
\usepackage{amsmath,amssymb} 
\usepackage{color}
\usepackage{epsfig}
\usepackage{graphicx}
\usepackage{array}
\usepackage{tabularx}
\usepackage{threeparttable}
\usepackage{subfigure}
\usepackage{url}

\usepackage[width=122mm,left=12mm,paperwidth=146mm,height=193mm,top=12mm,paperheight=217mm]{geometry}
\usepackage[pagebackref=false,breaklinks=true,colorlinks,bookmarks=false]{hyperref}

\newcolumntype{P}[1]{>{\centering\arraybackslash}p{#1}}
\newcommand{\vs}[1]{\textit{v.s. }}
\newcommand{\eg}[1]{\textit{eg}}
\newcommand{\ie}[1]{\textit{i.e. }}

\begin{document}
\pagestyle{headings}
\mainmatter
\def\ECCVSubNumber{2716}  

\title{Dynamic Dual-Attentive Aggregation Learning for Visible-Infrared Person Re-Identification} 

\titlerunning{Dynamic Dual-Attentive Aggregation for VI-ReID}
%
\author{Mang Ye\inst{1} \and
Jianbing Shen\inst{1}\thanks{Corresponding Author: \textit{Jianbing Shen}}\and
David J. Crandall \inst{2}\and
Ling Shao \inst{1,4}\and
Jiebo Luo \inst{3}}
\authorrunning{M. Ye et al.}
%
\institute{Inception Institute of Artificial Intelligence, UAE \and
Indiana University, USA \quad $^{3}$University of Rochester, USA\\
\quad $^4$ Mohamed bin Zayed University of Artificial Intelligence, UAE
\url{https://github.com/mangye16/DDAG}
}
\maketitle

\begin{abstract}
Visible-infrared person re-identification (VI-ReID) is a challenging cross-modality pedestrian retrieval problem. Due to the large intra-class variations and cross-modality discrepancy with large amount of sample noise, it is difficult to learn discriminative part features. Existing VI-ReID methods instead tend to learn global representations, which have limited discriminability and weak robustness to noisy images. In this paper, we propose a novel dynamic dual-attentive aggregation (DDAG) learning method by mining both intra-modality part-level and cross-modality graph-level contextual cues for VI-ReID. We propose an intra-modality weighted-part attention module to extract discriminative part-aggregated features, by imposing the domain knowledge on the part relationship mining. To enhance robustness against noisy samples, we introduce cross-modality graph structured attention to reinforce the representation with the contextual relations across the two modalities. We also develop a parameter-free dynamic dual aggregation learning strategy to adaptively integrate the two components in a progressive joint training manner. Extensive experiments demonstrate that DDAG outperforms the state-of-the-art methods under various settings.
\end{abstract}

\section{Introduction}
Person re-identification (Re-ID) techniques \cite{arxiv20reidsurvey,iccv15zheng} have achieved human-level performance with part-level deep feature learning \cite{iccv19attention,eccv18pcb,cvpr19multi}. However, most of these techniques consider images of people collected by visible-spectrum cameras in the daytime,
and thus are not applicable to night-time applications. Infrared cameras can be
used to collect imagery in low light conditions \cite{iccv17cross}, but matching this imagery to visible-spectrum images is a significant challenge.

\begin{figure}[t]
\centering
  \includegraphics[height=3.6cm]{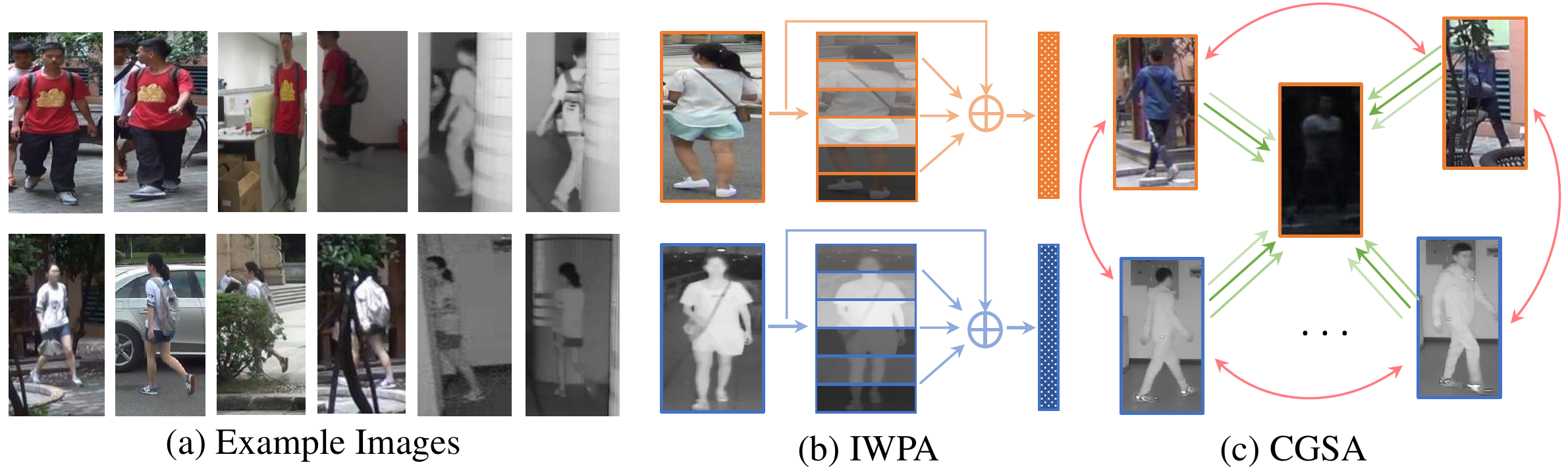}
  \caption{\small{Idea Illustration: (a) Example images from SYSU-MM01 dataset \cite{iccv17cross} with high sample noise due to data annotation/collection difficulty. Main components: (b) intra-modality weighted-part aggregation (IWPA), it learns discriminative part-aggregated features by mining the contextual part information within each modality. (c) cross-modality graph structured attention (CGSA), it enhances the representation by incorporating the neighborhood information from the two modalities.}}
\label{fig:idea}
\end{figure}

Cross-modality visible-infrared person re-identification (VI-ReID)~\cite{iccv17cross,tifs19vtreid} aims to solve this problem by matching images of people captured by visible and infrared (including near-~\cite{iccv17cross} and far-infrared (thermal)~\cite{sensors17}) cameras. VI-ReID is challenging due to large visual differences between the two modalities and changing camera environments, leading to large intra- and cross-modality variations. Moreover, due to difficulties in data collection and annotation, VI-ReID usually suffers from high sample noise caused by inaccurate person detection results, \eg, extreme background clutter, as shown in Fig.~\ref{fig:idea} (a). Related cross-modality matching studies have been extensively conducted in visible near-infrared (VIS-NIR) face recognition \cite{tifs19nir,aaai18nir}. However, the visual differences between images of people are much greater than those between face images, so those methods are not applicable for VI-ReID \cite{iccv17cross}.

These challenges make it difficult to reliably learn discriminative part-level features using state-of-the-art
single-modality Re-ID systems~\cite{eccv18pcb,mm18mgn,tip19part,cvpr19multi}. As a compromise, existing VI-ReID methods mainly focus on learning multi-modal sharable global features, either via one- \cite{ijcai18cross,cvpr19ivreid,iccv17cross} or two-stream networks \cite{tip19msr,tifs19vtreid}. Some work also integrates modality discriminant supervision \cite{ijcai18cross,tip19msr} or GAN generated images \cite{iccv19ivreid,cvpr19ivreid} to handle the modality discrepancy. However, global feature learning methods are
sensitive to background clutter and can not explicitly handle the modality discrepancy. In addition, part-based feature-learning
methods~\cite{eccv18pcb,mm18mgn,iccv17part,cvpr19multi} for single-modality Re-ID are typically incapable of capturing reliable
part features under a large cross-domain gap~\cite{iccv17cross}. Moreover, the learning is easily contaminated by noisy samples and destabilized when the appearance discrepancy is large across the two modalities. All of these challenges result in less discriminative cross-modality features and unstable training.

To address the above limitations, we propose a novel dynamic dual-attentive aggregation (DDAG) learning method with a two-stream network. DDAG includes two main components, as shown in Fig.~\ref{fig:idea}: an intra-modality weighted-part aggregation (IWPA) and a cross-modality graph structured attention (CGSA). Our main idea is to mine contextual cues at both an intra-modality part-level and cross-modality graph-level, to enhance feature representation learning. IWPA aims to learn discriminative part-aggregated features by simultaneously mining the contextual relations among the body parts within each modality and imposing the domain knowledge to handle the modality discrepancy. Our design is computationally efficient because we learn the modality-specific part-level attention rather than pixel-level attention \cite{cvpr18nonlocal,iclr19resatt}, and it also results in stronger robustness against background clutter. We further develop a residual BatchNorm connection with weighted-part aggregation to reduce the impact of noisy body parts, and adpatively handle the part discrepancy in the aggregated features.

CGSA focuses on learning an enhanced node feature representation by incorporating the relationship between the person images across the two modalities. We eliminate the negative impact of samples with large variations by exploiting the contextual information in the cross-modality graph, assigning adaptive weights to both intra- and cross-modality neighbors with a mutli-head attentive graph scheme \cite{nips17attend}. This strategy also reduces the modality discrepancy and smooths the training process. In addition, we introduce a parameter-free dynamic dual aggregation learning strategy to dynamically aggregate the two attentive modules in a mutli-task end-to-end learning manner, which enables complex dual-attentive network to converge stably, while simultaneously reinforcing each attentive component.
Our main contributions are as follows:
 \begin{itemize}
\item We propose a novel dynamic dual-attentive aggregation learning
  method to mine contextual information at both intra-modality
  part and cross-modality graph levels to facilitate feature
  learning for VI-ReID.
\item We design an intra-modality weighted-part attention module to
  learn discriminative part-aggregated representation, adaptively assigning the
  weights of different body parts.
\item We introduce a cross-modality graph structured attention scheme to enhance feature representations by mining the graphical relations between the person images across the two modalities, which smooths the training process and reduces the modality gap.
\item We establish a new baseline on two VI-ReID datasets, outperforming the state-of-the-art by a large margin.
 \end{itemize}

\section{Related Work}\label{sec:related}
\textbf{Single-Modality Person Re-ID} aims to match person images from visible
cameras~\cite{tcsvt19survey}. Existing works have achieved human-level performance on the widely-used datasets with end-to-end deep learning  \cite{cvpr19fusion,cvpr19interact,cvpr19occlusion,cvpr18att,cvpr19part,cvpr19rich}, either by global \cite{cvpr18att,arxiv17align} or part-level feature learning \cite{eccv18pcb,cvpr19part,cvpr19multi}. However, these approaches are usually unable to handle the ambiguous modality discrepancy in VI-ReID \cite{iccv17cross}, which limits their applicability in night-time surveillance scenarios.

\noindent \textbf{Cross-Modality Person Re-ID}
addresses person re-identification across different types of images, such as between
visible-spectrum and infrared~\cite{cvpr19ivreid,iccv17cross,aaai18vtreid}, varying illuminations \cite{tmm20illumination} or even between
images and non-visual data like text descriptions~\cite{eccv18text,iccv17nlp}.
For visible-Infrared-ReID (VI-ReID), Wu \textit{et al.} \cite{iccv17cross} introduced a zero-padding strategy with a one-stream network for cross-modality feature representation learning. A two-stream network with dual-constrained top-ranking loss was proposed in \cite{tifs19vtreid} to handle both the intra- and cross-modality variations. In addition, Dai~\textit{et al.} \cite{ijcai18cross} proposed an adversarial training framework with the triplet loss to jointly discriminate the identity and modality. Recently, Wang \textit{et al.} \cite{cvpr19ivreid} presented a dual-level discrepancy method with GAN to handle the modality difference at various levels. Similar technique was also adopted in \cite{iccv19ivreid}. Two modality-specific \cite{tip19msr} and modality-aware learning \cite{tip20mace} methods were proposed to handle the modality discrepancy at the classifier level. Meanwhile, other papers have investigated a better loss function \cite{arxiv19efficient,arxiv19hpiln} to handle the modality gap. However, these methods usually focus on learning global feature representations, which ignore the underlying relationship between different body parts and neighborhoods across two modalities.

Contemporaneously, some recent methods investigate the modality-aware collaborative ensemble learning \cite{tip20mace} or grayscale augmented tri-modal learning \cite{tifs20gray}. An intermediate X-modality is designed in \cite{aaai20xmodal} to address the modality discrepancy. A powerful baseline with non-local attention is presented in \cite{arxiv20reidsurvey}.

\noindent\textbf{Visible Near-Infrared Face Recognition} addresses the cross-modality face recognition problem, which is also closely related to
VI-ReID~\cite{aaai17nir,tifs19nir,tip19nirvis,tip17face,aaai18nir,tnnls20pang}. Early research mainly focused on learning modality-aware metrics \cite{tifs19pang} or dictionaries \cite{iccv13scdl}. With the emergence of deep neural networks, most methods now focus on learning multi-modal sharable features \cite{aaai18nir}, cross-modality matching models \cite{ijcv17deep} or disentangled representations \cite{aaai19nir}. However, the modality difference of VI-ReID is much greater than that of face recognition due to the different camera environments and large visual appearances change, which limits the applicability of their methods to the VI-ReID \cite{aaai18vtreid,ijcai20survey}.

\noindent\textbf{Attention Mechanisms} have been widely used in various
applications to enhance the feature representation \cite{cvpr18dualatt,nips17attend,icml15attend,iccv19cao}. For person Re-ID, attention is used to combine the spatial-temporal information from different video frames \cite{iccv19biattention,iccv19secatten,cvpr18diversity,bmvc19atten}. Some work \cite{cvpr18hacnn,iccv17Attentive,cvpr19aanet} has also investigated using multi-scale or different convolutional channels to capture the pixel-level/small-region-level attentions \cite{cvpr19shao,tifs18shao}. However, they are usually unstable for optimization in VI-ReID due to the large cross-modality discrepancy and noise.

Our part-attention module is also closely related to non-local networks \cite{cvpr18nonlocal,iclr19resatt}. However, the pixel-level design of these models is sensitive and inefficient for handling the noise encountered in VI-ReID task. In comparison, we design a learnable weighted part-level attention with a BatchNorm residual connection.
\begin{figure*}[t]
  \centering
  \includegraphics[height = 5cm]{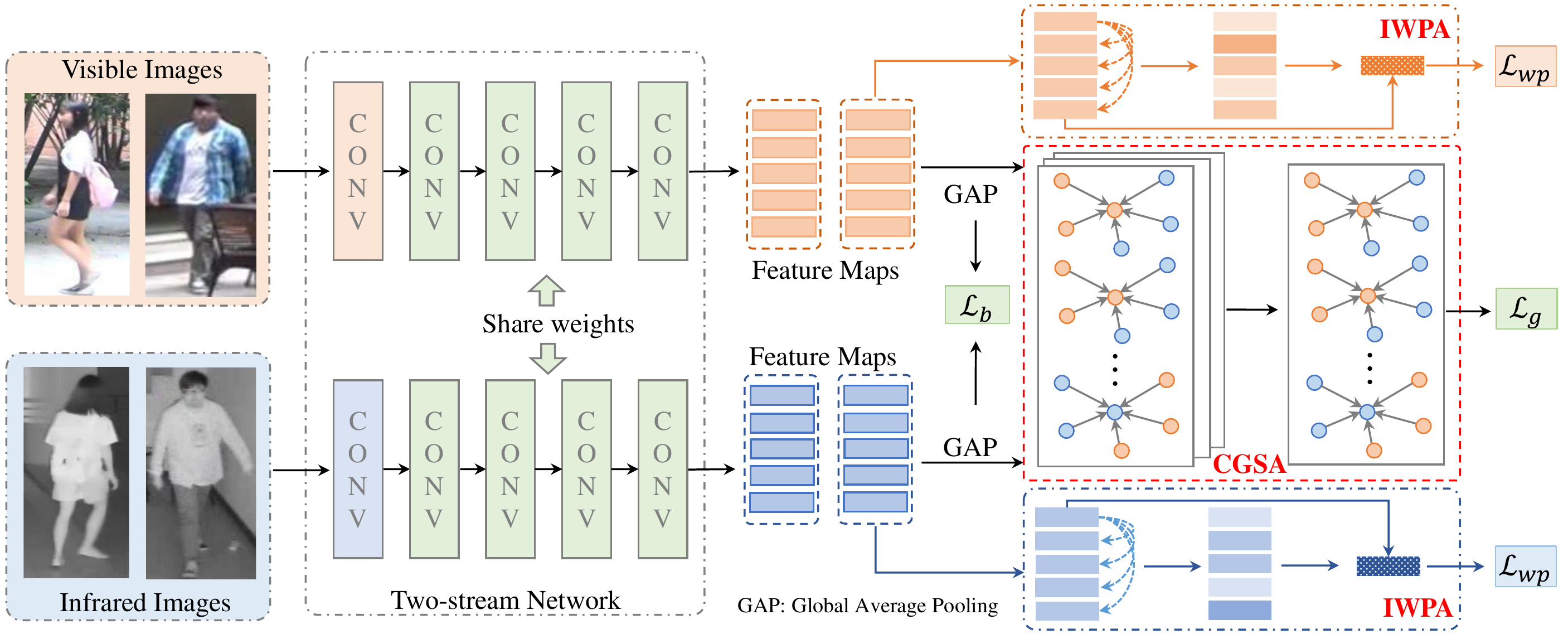}\\
  \caption{\small{The proposed DDAG learning framework for VI-ReID. \textbf{IWPA}: Intra-modality Weighted-Part Aggregation for discriminative part-aggregated features learning by mining the contextual part relations within each modality. \textbf{CGSA}: Cross-modality Graph Structured Attention for global feature learning by utilizing the neighborhood structural relations across two modalities. We further introduce a parameter-free dynamic dual aggregation learning strategy to adaptively aggregate two components.} }\label{fig:framework}
\end{figure*}

\section{Proposed Method}
Fig.~\ref{fig:framework} provides an overview of our proposed dynamic
dual-attentive aggregation learning (DDAG) method. DDAG is developed on top of a two-stream network (\S \ref{sec:base}), and contains an intra-modality weighted-part attention for discriminative part-aggregated features learning (\S \ref{sec:iwpa}) and a cross-modality graph structured attention for shared global feature learning (\S \ref{sec:cgsa}). Finally, we propose a parameter-free dynamic dual aggregation learning strategy to adaptively aggregate the two components for end-to-end joint training (\S \ref{sec:dual}).

\subsection{Baseline Cross-Modality Re-ID}\label{sec:base}
We first present our baseline cross-modality Re-ID model with a two-stream network for incorporating two different modalities. To handle the different properties of the two modalities, the network parameters of
the first convolutional block\footnote{We adopt ResNet50 as the backbone network, following \cite{iccv19ivreid,cvpr19ivreid,tifs19vtreid}.} in each stream are
different in order to capture modality-specific low-level feature
patterns. Meanwhile, the network parameters of the deep convolutional blocks are shared for two modalities in order to learn modality-sharable middle-level feature representations. After the convolutional layers with adaptive pooling, a shared batch normalization layer is added to learn the shared feature embedding. Compared with the two-stream structures in \cite{aaai19hsme,arxiv19haijun,tifs19vtreid,tip20mace}, our design captures more discriminative features by mining sharable information in middle-level convolutional blocks rather than high-level embedding layers.

To learn discriminative features, we combine the identity loss $\mathcal{L}_{id}$ and online hard-mining triplet loss $\mathcal{L}_{tri}$ \cite{pami20isif} as our baseline learning objective $\mathcal{L}_{b}$,
\begin{equation}\label{eq:base}
\mathcal{L}_{b}= \mathcal{L}_{id} + \mathcal{L}_{tri}.
\end{equation}
The identity loss $\mathcal{L}_{id}$ encourages an identity-invariant feature representation. The triplet loss $\mathcal{L}_{tri}$ optimizes the triplet-wise relationships among different person images across the two modalities.

\begin{figure}[t]
\centering
   \includegraphics[height=5cm]{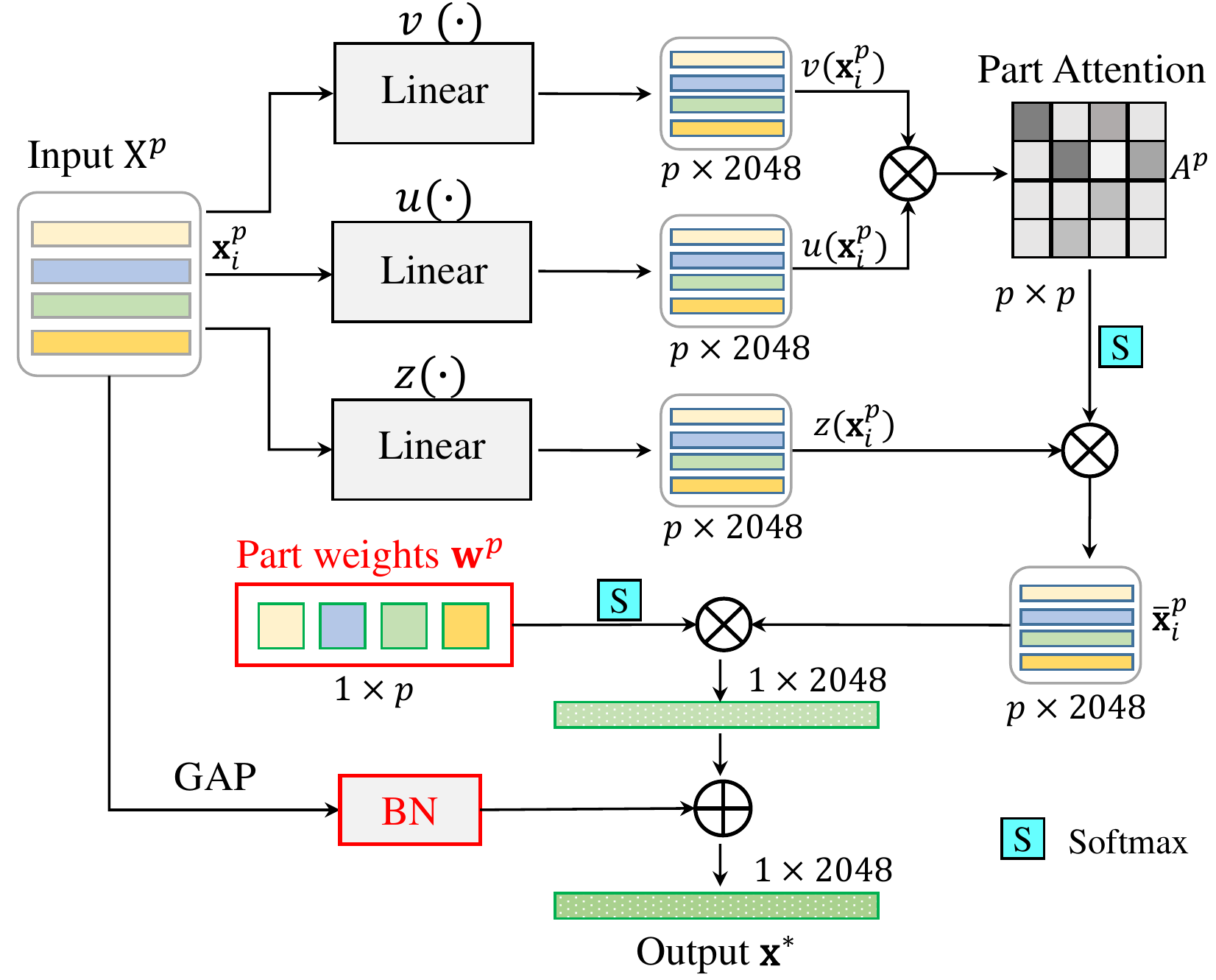}
   \caption{\small{Illustration of the proposed IWPA module, which mines the part-level relationships to learn the weighted-part aggregation with a residual BatchNorm connection. }}
\label{fig:mwpa}
\end{figure}
\subsection{Intra-modality Weighted-Part Aggregation} \label{sec:iwpa}
As an alternative to the global feature learning in existing VI-ReID methods \cite{ijcai18cross,cvpr19ivreid,iccv17cross}, this subsection presents a novel part-agggregated feature learning method for VI-ReID, namely intra-modality weighted-part aggregation (IWPA, as shown in Fig.~\ref{fig:mwpa}). IWPA mines the contextual information in local parts to formulate an enhanced part-aggregated representation to address the complex challenges. It first learns the within-modality part attention with a modified non-local module, and then uses a learnable weighted-part aggregation strategy with residual BatchNorm (RBN) to stabilize and reinforce the training process.

\noindent\textbf{Part Attention.} The input of our IWPA module is the extracted feature maps from the last residual block of the network, from which we extract the attention-enhanced part features. We denote the output feature maps of the last convolutional block as $\{X = \mathbf{x}_k \in\mathbb{R}^{ C \times H \times W  }\}_{k=1}^K$, where $C$ represents the channel dimension ($C=2048$ in our experiments), $H$ and $W$ represent the feature map size, and $K$ represents the batch size. To obtain the part features, the feature maps are directly divided into $p$ non-overlapping parts with a region pooling strategy. The part features of each input image are then represented by $X^p = \{\mathbf{x}^p_i\in \mathbb{R}^{ C \times 1}\}_{i=1}^{p}$. Similar to \cite{cvpr18nonlocal}, we feed each part into three $1\times1$ convolutional layers $u(\cdot)$, $v(\cdot)$ and $z(\cdot)$. The intra-modality part-based non-local attention $\alpha^p_{i,j} \in[0,1]^{p \times p} $ is then
\begin{equation}\label{eq:attp}
\alpha^p_{i,j}= \frac{f(\mathbf{x}^p_i,\mathbf{x}^p_j)}{\sum_{\forall{j}}f(\mathbf{x}^p_i,\mathbf{x}^p_j)},
\end{equation}
where $f(\mathbf{x}^p_i,\mathbf{x}^p_j)$ represents the pairwise similarity between two part features. To enhance the discriminability, an exponential function is added to magnify the relationship, which enlarges the part attention discrepancy \cite{arxiv18heat}. It is formulated by
\begin{equation}\label{eq:gaus}
f(\mathbf{x}^p_i,\mathbf{x}^p_j) = \exp (u(\mathbf{x}^p_i)^T v(\mathbf{x}^p_j)),
\end{equation}
where $u(\mathbf{x}^p_i) = W_u \mathbf{x}^p_i$ and $v(\mathbf{x}^p_j) = W_v \mathbf{x}^p_j$ are two embeddings with convolutional operations $u(\cdot)$ and $v(\cdot)$. $W_u$ and $W_v$ are the corresponding weight parameters in $u$ and $v$. With the exponential function, our attention calculation can be treated as a normalization with a softmax function. Note that our attention map is $p\times p$ to capture the part relationships, which is much smaller than that of pixel-level attention $HW \times HW$ in \cite{cvpr18nonlocal,iclr19resatt}, making it more efficient. Meanwhile, the part relation is robust against noisy regions and local clutters in the person images.

With the learned part attention, attention-enhanced part features are then represented by the inner product of the embedded part features $ z(\mathbf{x}^p_i)$ and the calculated attention $A^p$, which is formulated by
\begin{equation}\label{eq:part}
\bar{\mathbf{x}}_i^p = \mathbf{a}^p_i * z(\mathbf{x}^p_i),
\end{equation}
where $\mathbf{a}^p_i \in A^p= \{\alpha_{i,j}^p\} ^{p \times p}$ is the calculated part attention map.
Therefore, the refined part features consider the relationship between different body parts. However, the simple average pooling or concatenation of part features is not powerful enough for fine-grained person Re-ID task, and may cause noisy parts accumulation. Meanwhile, it is inefficient to train multiple part-level classifiers, as in \cite{eccv18pcb,tip20mace}. To address these issues, we design a residual BatchNorm (RBN) weighted-part aggregation strategy.

\noindent\textbf{Residual BatchNorm Weighted-part Aggregation.} This idea consists of two main parts: First, we use a residual BatchNorm connection of the original input feature map $\mathbf{x}^o$ after average pooling, and the residual learning strategy enables very deep neural networks to be trained and stabilizes the training process. Second, we use a learnable weighted combination of attention-enhanced part features to formulate a discriminative part-aggregated feature representation. In summary, it is formulated by
\begin{equation}\label{eq:respart}
\mathbf{x}^* = \mathrm{BN}(\mathbf{x}^o) + \sum\nolimits_{i=1}^p w^p_i \mathbf{\bar{x}}^p_i,
\end{equation}
where $\mathbf{x}^o \in  \mathbb{R}^{ C \times 1} $ represents the global adaptive pooling output of the input feature map $X^p$. $\mathrm{BN}$ indicates the batch normalization operation, and $\mathbf{w}^p = \{w^p_i\}_{i=1}^{p}$ represents a learnable weight vector of different parts to handle the modality discrepancy. Our design has three primary advantages: (1) it avoids multiple part-level classifier learning \cite{eccv18pcb}, making it computationally efficient for both training and testing, and it is more robust to background clutter compared to the pixel-level attention techniques \cite{cvpr18hacnn,cvpr18nonlocal}; (2) it enhances the discrimination power by adaptively aggregating attentive part features in the final feature representation; and (3) the residual BatchNorm (RBN) connection performs much better than the widely-used general residual connection with identity mapping \cite{cvpr16resnet,iclr19resatt} (as verified in \S \ref{sec:self}), stabilizing the training process and enhancing the representational power for the cross-modality Re-ID under abundant noise. We use $\mathbf{x}^*$ as the representation of an input sample in the testing phase.

\subsection{Cross-modality Graph Structured Attention}\label{sec:cgsa}
Another major challenge is that VI-ReID datasets often contain many incorrectly annotated images or image pairs with large visual differences across the two modalities (as shown in Fig. \ref{fig:idea}), making it difficult to mine the discriminative local part features and damaging the optimization process. In this subsection, we present our cross-modality graph structured attention, which incorporates the structural relations across two modalities to reinforce the feature representations. The main idea is that the feature representations of person images belonging to the same identity across the two modalities are mutually beneficial.

\noindent\textbf{Graph Construction.} At each training step, we adopt an identity-balanced sampling strategy for training \cite{tifs19vtreid}. Specifically, for each of $n$ different
randomly-selected identities, $m$ visible and $m$ infrared images are
randomly sampled, resulting in $K = 2mn$ images in each
training batch. We formulate an undirected graph $\mathcal{G}$ with a normalized adjacency matrix,
\begin{equation}\label{eq:graph}
A^g = A^{g}_0 + \mathbb{I}_K,
\end{equation}
where $A^{g}_0(i,j) = l_i * l_j$ ($l_i$ and $l_j$ are the corresponding one-hot labels of two graph nodes). $\mathbb{I}_K$ is an identity matrix, indicating that each node is connected to itself. The graph construction is efficiently computed by matrix multiplication between the one-hot labels in each training batch.

\noindent\textbf{Graph Attention.} This measures the importance of a node $i$ to another node $j$ within the graph, across two modalities. We denote the input node features by $X^o = {\{\mathbf{x}^o_k \in \mathbb{R}^{ C \times 1}\}}_{k=1}^{K}$, which are outputs of the pooling layer. The graph attention coefficients $\alpha^g_{i,j}\in[0,1]^{K \times K}$ are then computed by
\begin{equation}\label{eq:attg}
\alpha^g_{i,j}= \frac{\exp (\Gamma( \lceil h(\mathbf{x}^o_i),h(\mathbf{x}^o_j)\rfloor \cdot \mathbf{w}^g))}{\sum_{\forall{A^{g}(i,k)>0}}\exp (\Gamma( \lceil h(\mathbf{x}^o_i),h(\mathbf{x}^o_k)\rfloor \cdot  \mathbf{w}^g))},
\end{equation}
where $\Gamma(\cdot)$ represents the $\mathrm{LeakyRelu}$ operation. $\lceil, \rfloor$ is the concatenation operation. $h(\cdot)$ is a transformation matrix to reduce the input node feature dimension $C$ to $d$, and $d$ is set to 256 in our experiments. $\mathbf{w}^g \in \mathbb{R}^{ 2d \times 1 }$ represents a learnable weighting vector that measures the importance of different feature dimensions in the concatenated features, similar to \cite{iclr18gat}. Note that our design fully utilizes relations between all the images across two modalities, reinforcing the representation using context information of the same identity.

To enhance the discriminability and stabilize the graph attention learning, we employ a multi-head attention technique \cite{nips18collaborative} by learning multiple $h^l(\cdot)$ and $\mathbf{w}^{l,g}$ ($l = 1,2\cdots,L$, $L$ is the total number of heads) with the same structure and optimizing them separately. After concatenating the outputs of multiple heads, the graph structured attention-enhanced feature is then represented by
\begin{equation}\label{eq:head}
\mathbf{x}^g_i = \phi{\lceil \sum\nolimits_{\forall{A^{g}(i,k)>0}}\alpha^{g,l}_{i,j} \cdot h^l(\mathbf{x}^o_j)\rfloor}_{l=1}^L,
\end{equation}
and $\mathbf{x}^g_i$ is then robust to outlier samples, where $\phi$ is the $\mathrm{ELU}$ activation function. To guide the cross-modality graph structured attention learning, we introduce another graph attention layer with a one-head structure, where the final output node features are represented by $X^{g'} = {\{\mathbf{x}^{g'}_i\}}_{k=1}^K$. We adopt the negative log-likelihood (NLL) loss function for the graph attention learning, formulated by
\begin{equation}\label{eq:lg}
\mathcal{L}_g = - \sum\nolimits_i^{K} \log(\mathrm{softmax}(\mathbf{x}^{g'}_i)).
\end{equation}
\subsection{Dynamic Dual Aggregation Learning}\label{sec:dual}
Incorporating the above proposed intra-modality weighted-part attention and cross-modality graph-structured attention into an end-to-end joint learning framework is highly challenging. This is primarily because the two components focus on different learning objectives with very deep network structures, and directly combining them easily will result in \textit{gradient explosion} problem after several steps. Moreover, the features from the same identity across two modalities are quite different in VI-ReID due to the large cross-modality variations, as demonstrated in Fig. \ref{fig:idea}. Therefore, the graph-structured attention would be unstable due to the large feature difference across the two modalities at the early stage.

\begin{figure}[t]
\centering
   \includegraphics[width=5cm]{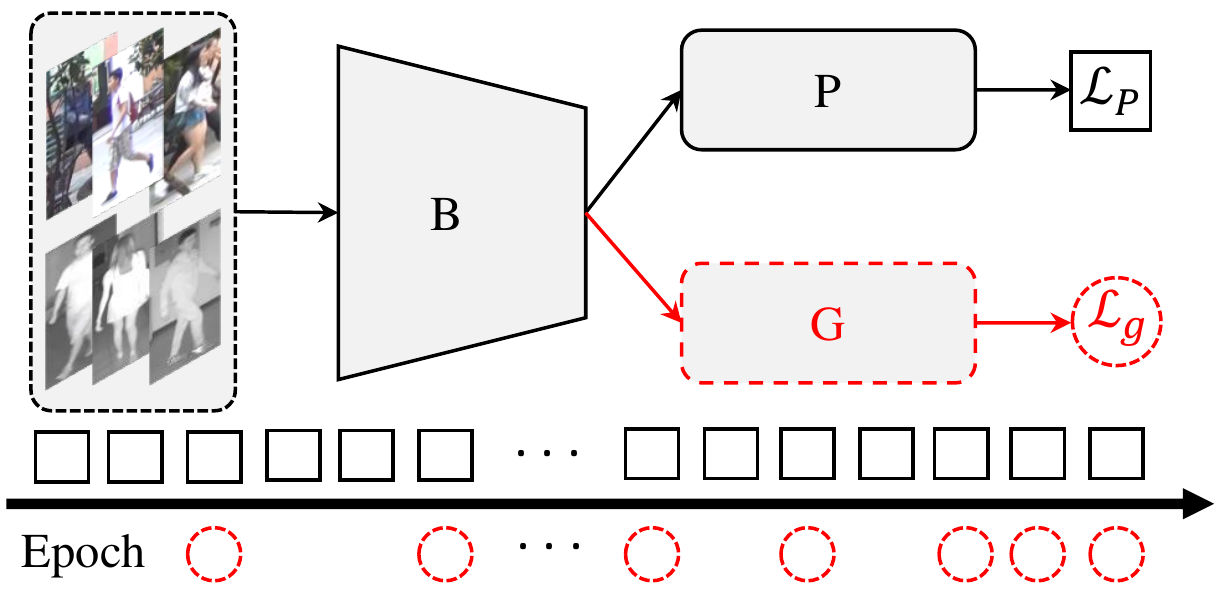}
   \caption{\small{Illustration of parameter-free dynamic dual aggregation learning. We decompose the overall training framework into two parts: instance-level part-aggregated feature learning $\mathcal{L}_P$ and graph-level global feature learning $\mathcal{L}_g$. We treat $\mathcal{L}_P$ as the dominant loss and progressively add $\mathcal{L}_g$ in the overall training process.  }}
\label{fig:dualtrain}
\end{figure}
To address the above issues, we introduce a dynamic dual aggregation learning strategy to adaptively integrate the above introduced two components. Specifically, we decompose the overall framework into two different tasks, instance-level part-aggregated feature learning $\mathcal{L}_P$ and graph-level global feature learning $\mathcal{L}_g$. The instance-level part-aggregated feature learning $\mathcal{L}_P$ is a combination of the baseline learning objective $\mathcal{L}_{b}$ and the intra-modality weighted-part attention loss $\mathcal{L}_{wp}$, represented by
\begin{equation}\label{eq:Ploss}
\mathcal{L}_P = \mathcal{L}_{b}  \underbrace{ -\frac{1}{K}\sum\nolimits_{i=1}^K y_i \log(p(y_i|\mathbf{x}^*_i))}_{part\ attention\ loss \ \mathcal{L}_{wp}},
\end{equation}
where $p(y_i|\mathbf{x}^*_i)$ represents the probability of $\mathbf{x}^*_i$ being correctly classified into the groundtruth label $y_i$. The second term represents the instance-level part-aggregated feature learning with weighted-part attention within each modality. It is formulated by the identity loss on top of the aggregated part feature $\mathbf{x}^*$.

\noindent\textbf{Dynamic Dual Aggregation Learning.} Motivated by multi-task learning \cite{icml18gradnorm}, our basic idea is that the instance-level part-aggregated feature learning $\mathcal{L}_P$ acts as the dominant loss, and then we progressively add the graph-level global feature learning loss $\mathcal{L}_g$ for optimization. The main reason for doing this is that it is easier to learn an instance-level feature representation with $\mathcal{L}_P$ at an early stage. With a better learned network, the graph-level global feature learning optimizes the features using the relationship between the person images across the two modalities, denoted by
\begin{equation}\label{eq:finalloss}
\mathcal{L}^t = \mathcal{L}_P^t + \frac{1}{1 + \mathbb{E}(\mathcal{L}^{t-1}_P)}\mathcal{L}_g^t,
\end{equation}
where $t$ is the epoch number, and $\mathbb{E}(\mathcal{L}^{t-1}_P)$ represents the average loss value in the previous epoch. In this dynamic updating framework (as shown in Fig.~\ref{fig:dualtrain}), the graph-level global loss $\mathcal{L}_g$ is progressively added into the overall learning process. This strategy shares a similar spirit to the gradient normalization in multi-task learning \cite{icml18gradnorm}, but it does not introduce any additional hyper-parameter tuning.

When we optimize $\mathcal{L}_P$, the parameters of the identity classifier in the weighted-part attention loss $\mathcal{L}_{wp}$ are the same as those for the identity classifier in $\mathcal{L}_{b}$. Our motivation here is that this setting can guarantee that instance-level part-aggregated feature learning is directly performed on the part-aggregated features rather than additional classifiers, ensuring the discriminability of the learned features. Meanwhile, it avoids additional network parameters training.

\section{Experimental Results}
\subsection{Experimental Settings}
We use two publicly available VI-ReID datasets (SYSU-MM01 \cite{iccv17cross} and RegDB \cite{sensors17}) for the experiments. The rank-$k$ matching accuracy and mean Average Precision (mAP) are used as evaluation metrics, following \cite{iccv17cross}.

SYSU-MM01 \cite{iccv17cross} is a large-scale dataset collected by four RGB and two near-infrared cameras. The major challenge is that person images are captured in both indoor and outdoor environments. In total, the training set contains 22,258 visible and 11,909 near-infrared images of 395 identities. It contains two different testing settings, \textit{all-search} and \textit{indoor-search} mode. The query set contains 3,803 images of 96 identities captured from near-infrared cameras. The gallery set contains the images captured by all four RGB cameras in the \textit{all-search} mode, while the \textit{indoor-search} mode contains images of two indoor RGB cameras. Details on the experimental settings can be found in \cite{iccv17cross}.

RegDB \cite{sensors17} is collected by a dual-camera system, including one visible and one far-infrared camera. In total, this dataset contains 412 person identities, each of which has 10 visible and 10 far-infrared images. Following \cite{aaai18vtreid}, we randomly select 206 identities for training and the remaining 206 identities for testing. Thus the testing set contains 2,060 visible and 2,060 far-infrared images. We evaluate both visible-to-infrared and infrared-to-visible query settings. The performance is averaged over ten trials on random training/testing splits \cite{cvpr19ivreid,aaai18vtreid}.


\noindent\textbf{Implementation details.} Our proposed method is implemented in PyTorch. Following existing VI-ReID works, ResNet50 \cite{cvpr16resnet} is adopted as our backbone network for fair comparison, following \cite{arxiv20reidsurvey}. The first residual block is specific for each modality while the other four blocks are shared. The stride of the last convolutional block is set to 1, in order to obtain a fine-grained feature map. We initialize the convolutional blocks with the pre-trained ImageNet parameters, as done in \cite{tifs19vtreid}. All the input images are firstly resized to $288\times 144$. We adopt random cropping with zero-padding and horizontal flipping for data augmentation. SGD optimizer is adopted for optimization, and the momentum parameter is set to 0.9. We set the initial learning rate to 0.1 with a warm-up strategy \cite{arxiv19trick}. The learning rate decays by 0.1 at the 30th epoch and 0.01 at the 50th epoch, with a total of 80 training epochs. By default, we randomly select 8 identities, and then randomly select 4 visible and 4 infrared images to formulate a training batch. We set $p=3$ in Eq. \ref{eq:respart}, $L=4$ in Eq. \ref{eq:head}.

\begin{table*}[t]\scriptsize
\centering
\caption{\label{tab:component}Evaluation of each component on the large-scale SYSU-MM01 dataset. ``$B$" represents the baseline results with a two-stream network trained by $\mathcal{L}_{b}$. ``$P$" denotes the intra-modality weighted-part attention. ``$G$" indicates the cross-modality graph structured attention. Dynamic dual-learning is adopted when aggregating two components. Rank at $r$ accuracy(\%) and mAP (\%) are reported. }
\begin{tabular}{p{1.3cm}|P{0.95cm}P{0.95cm}P{0.95cm}P{0.95cm}P{0.95cm}||P{0.95cm}P{0.95cm}P{0.95cm}P{0.95cm}P{0.9cm}}\hline
Datasets   &\multicolumn{5}{c||}{\textit{ All Search}}  & \multicolumn{5}{c}{\textit{ Indoor Search}} \\\hline
  Methods       & $r=1$ & $r=5$ & $r=10$ & $r=20$ & mAP     & $r=1$   & $r=5$   & $r=10$    & $r=20$ & mAP\\\hline
  $B$                     & 48.18 & 75.81 & 85.73  & 93.52  & 47.64   & 49.52 & 78.86  & 88.70 & 95.27 & 58.12 \\
    $B + P$                 & 53.69 & 81.16 & 88.38 & 94.56   & 51.37  & 58.08 & 84.91  & 92.37  & 97.26 & 65.07 \\
  $B + G$                 & 50.75 & 78.43 & 86.71 & 93.62   & 49.73   & 52.90 & 83.50  & 92.65  & 97.75 & 62.26 \\
    $B + P + G$            & 54.75   &  82.31  & 90.39 & 95.81  & 53.02  & 61.02 & 87.13 & 94.06 & 98.41 & 67.98 \\\hline
 \end{tabular}
\end{table*}

\subsection{Ablation Study}\label{sec:self}
\textbf{Evaluation of Each Component.}
This subsection evaluates the effectiveness of each component on the SYSU-MM01 dataset under both \textit{all-search} and \textit{indoor search} modes. Specifically,
``$B$" represents the baseline results with a two-stream network trained by $\mathcal{L}_{b}$. ``$P$" denotes the intra-modality weighted-part aggregation. ``$G$" indicates the cross-modality graph structured attention.

 We make several observations through the results shown in Table \ref{tab:component}.
\textit{1) Effectiveness of baseline}: Using shared convolutional blocks, we achieve better performance than the two-stream network in \cite{tip19msr,arxiv19haijun,tip20mace,tifs19vtreid}. Meanwhile, some training tricks taken from single-modality Re-ID \cite{cvpr19multi} also contributes to this super baseline.
\textit{2) Effectiveness of $P$}: the intra-modality weighted-part aggregation significantly improves the performance. This experiment demonstrates that learning part-level weighted-attention features is beneficial for cross-modality Re-ID. \textit{3) Effectiveness of $G$}: When we include the cross-modality graph structured attention ($B+G$), performance is improved by using the relationship between the person images across two modalities to reduce the modality discrepancy. \textit{4) Effectiveness of dual-aggregation}: When aggregating two attention modules with the dynamic dual aggregation strategy, the performance is further improved, demonstrating that these two attentions are mutually beneficial to each other.
\begin{table}[h]\scriptsize
\centering
\caption{\label{tab:part}Evaluation of re-weighted part attention with different designs on the SYSU-MM01 dataset (\textit{all-search mode}). Rank at $r$ accuracy (\%) and mAP (\%) are reported. (\textit{Setting: Baseline + Part attention.)}}
\begin{tabular}{p{1.4cm}|p{1.4cm}|P{1.4cm}P{1.4cm}P{1.4cm}|P{1.4cm}}
  \hline
  Method &  Res.  & $r=1$   & $r=10 $   & $r=20$   & mAP  \\\hline
  B        & N/A  & 48.18  & 85.73  & 93.52  & 47.64 \\
  avg & Res      & 48.34  & 86.03 & 93.72   & 48.43      \\
  concat & Res    & 50.34  & 86.43 & 94.19   & 49.77      \\
  weight & Res & 51.06  & 86.78 & 94.39   & 49.92\\\hline
  weight & RBN &  53.69  & 88.38 & 94.56   & 51.37 \\\hline
 \end{tabular}
\end{table}

\begin{table}\scriptsize
\centering
\caption{\label{tab:graph}Evaluation of graph attention on the SYSU-MM01 dataset  (\textit{all-search mode}). $N_g$ represents the number of images selected for graph construction. Rank at $r$ accuracy (\%) and mAP (\%) are reported. (\textit{Setting: Baseline + Graph attention.)}}
\begin{tabular}{l|P{0.95cm}P{0.95cm}P{0.95cm}P{0.95cm}P{0.95cm}}
  \hline
  $N_g$     & 0 & 1  & 2   & 4  & 8  \\\hline
  Rank-1   & 48.18  & 49.26   &49.85  &  50.45 & 50.75       \\
  mAP    & 47.64 & 48.42 & 49.12  & 49.46    & 49.73      \\\hline
 \end{tabular}
\end{table}

\noindent\textbf{Why Weighted Part Attention with RBN?}
We next compare different part attention designs on the SYSU-MM01 dataset under the \textit{all-search} mode. The results are shown in Table \ref{tab:part} and we make several observations. (1) \textit{Effectiveness of weighted scheme}. We compare the \textit{weighted} part features with average/concatenation schemes (termed as \textit{weighted}, \textit{avg} and \textit{concat} in Table  \ref{tab:part}). We observe that the proposed learnable weighted-part scheme performs consistently better than its two counterparts. Another benefit of the weighted aggregation is that the feature dimension of final representation is much smaller than the concatenation strategy in \cite{eccv18pcb}, which is more suitable for real applications with resource-demanding scenarios. (2) \textit{Effectiveness of residual BN (RBN) scheme}. We compare the general residual connection with the residual BN connection. Results demonstrate that RBN performs significantly better than the general residual connection. This suggests that the BN operation enhances the predictive and stable behavior of the training process \cite{nip18bn}, which is more suitable for VI-ReID with abundant noise. Note that the performance significantly drops without the residual connection.

\noindent\textbf{Why Graph Structured Attention?}
We now evaluate the effect of different numbers ($N_g$) of selected images for graph attention calculation. The results are shown in Table \ref{tab:graph}. A larger $N_g$ means that more neighbor images from the same identity are considered and the relationship is more reliable. Thus the accuracy is consistently improved with increasing $N_g$, demonstrating that the graph structured attention can largely reduce the modality discrepancy. Moreover, the infrared images capture less information than the visible images, but with much more noise. Mining the relation across two modalities, especially from the visible images, is thus beneficial for the cross-modality feature learning. The graph attention might also be applied in single-modality person re-identification.

\begin{figure}[t]
\begin{minipage}{0.49\linewidth}
\centering
 \includegraphics[width=0.69\linewidth,height = 4cm]{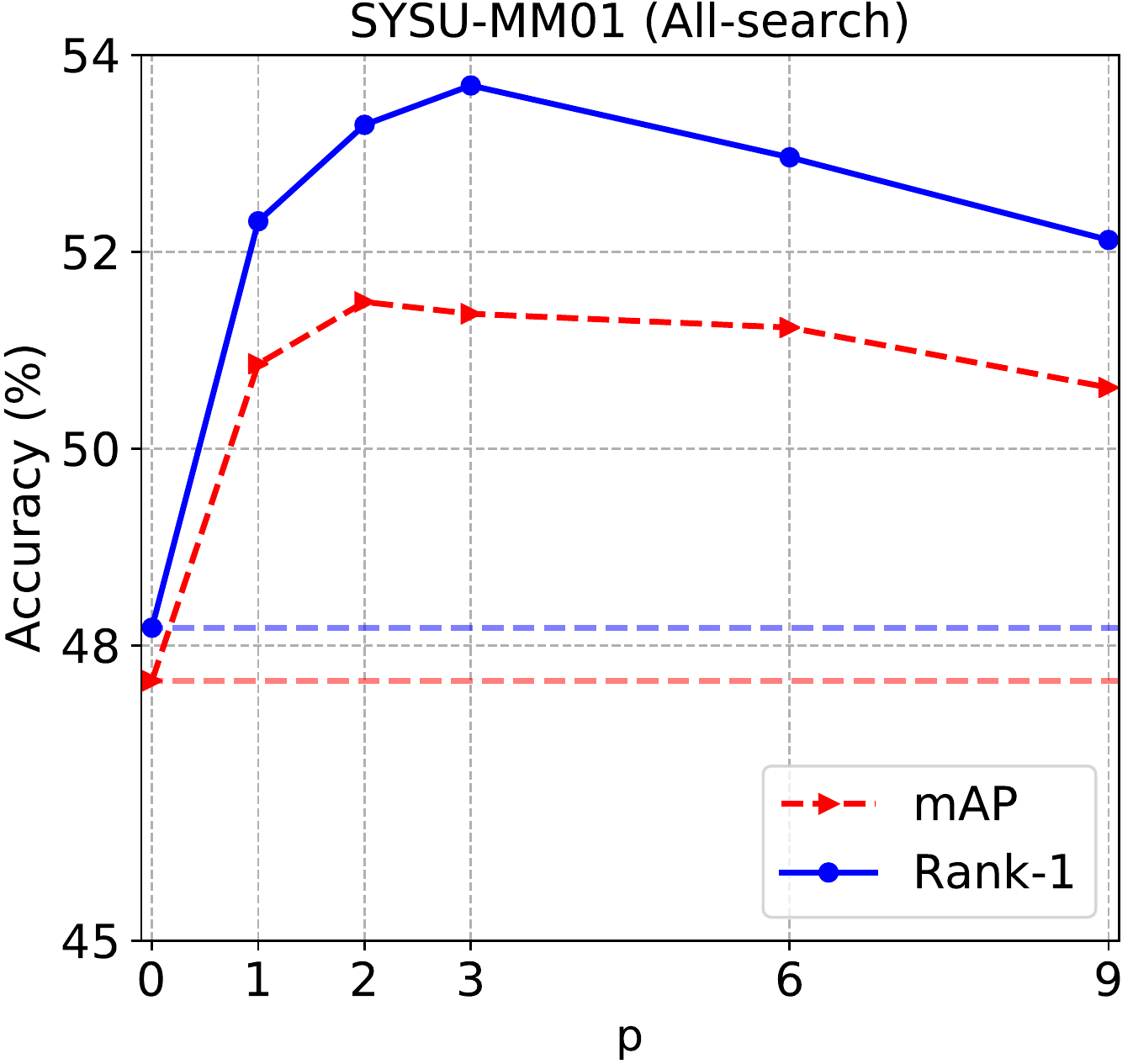}\\
\end{minipage}
\begin{minipage}{0.49\linewidth}
\centering
 \includegraphics[width=0.69\linewidth,height = 4cm]{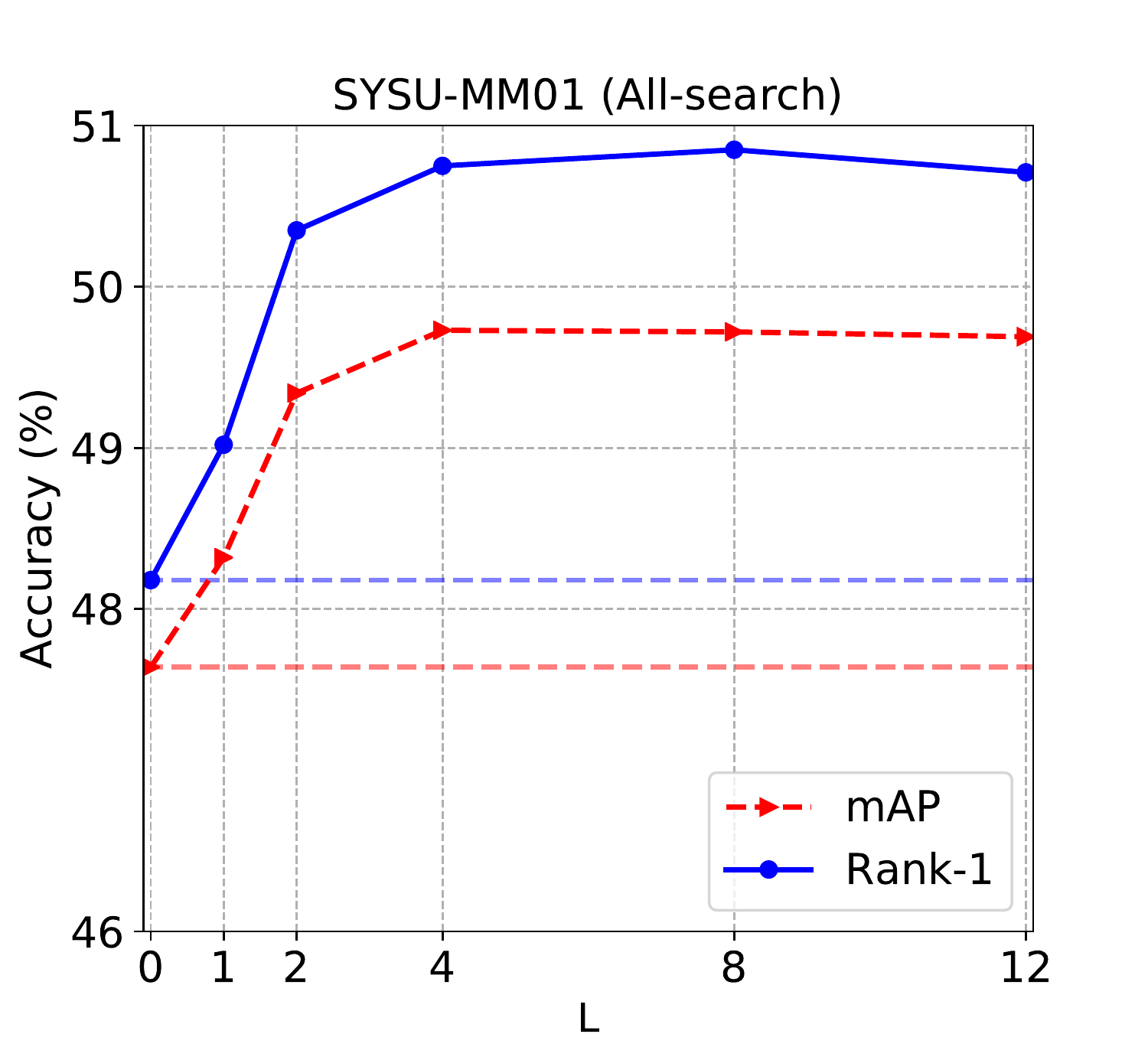}\\
\end{minipage}
\caption{\small{Evaluation of different body parts $p$ in Eq.~\ref{eq:part} (left) and different numbers of graph attention heads $L$ in Eq.~\ref{eq:head} (right) on SYSU-MM01 dataset, under the challenging \textit{all-search} mode. Rank-1 matching accuracy (\%) and mAP (\%) are reported.\label{fig:param} }}
\end{figure}

\begin{table*}[t]\scriptsize
\centering
 \caption{\label{tab:sysu}\small{Comparison with the state-of-the-arts on SYSU-MM01 dataset on two different settings. Rank at $r$ accuracy (\%) and mAP (\%) are reported. }}
\begin{tabular}{p{2cm}|p{1.0cm}|p{1.0cm}<{\centering}p{1.0cm}<{\centering}p{1.0cm}<{\centering}|p{1.0cm}<{\centering}||p{1.0cm}<{\centering}p{1.0cm}<{\centering}p{1.0cm}<{\centering}|p{1.0cm}<{\centering}}
\hline
\multicolumn{2}{c|}{Settings} &\multicolumn{4}{c||}{\textit{All Search}}  & \multicolumn{4}{c}{\textit{Indoor Search}} \\\hline
Method   &Venue    & $r=1$     & $r=10$    & $r=20$ & mAP            & $r=1$     & $r=10$    & $r=20$ & mAP\\\hline
  One-stream \cite{iccv17cross} &\scriptsize{ICCV17}& 12.04 & 49.68   & 66.74       & 13.67  & 16.94 & 63.55   & 82.10     & 22.95     \\
  Two-stream \cite{iccv17cross} &\scriptsize{ICCV17}& 11.65 & 47.99  & 65.50       & 12.85  & 15.60 & 61.18   & 81.02     & 21.49     \\
  Zero-Pad \cite{iccv17cross} &\scriptsize{ICCV17}&14.80 &54.12   &71.33       &15.95  & 20.58 & 68.38   & 85.79     & 26.92     \\
  TONE \cite{aaai18vtreid} &\scriptsize{AAAI18}&12.52 & 50.72   &68.60      & 14.42           & 20.82 & 68.86  & 84.46     & 26.38      \\
  HCML \cite{aaai18vtreid}  &\scriptsize{AAAI18}&14.32 &53.16   &69.17  & 16.16 & 24.52  & 73.25  & 86.73     & 30.08    \\
  cmGAN \cite{ijcai18cross} &\scriptsize{IJCAI18}& 26.97   & 67.51 & 80.56 & 31.49 & 31.63 & 77.23   & 89.18    & 42.19   \\
  BDTR \cite{tifs19vtreid}  &\scriptsize{IJCAI18}& 27.32 & 66.96 & 81.07 & 27.32 & 31.92 & 77.18 & 89.28 & 41.86  \\
  eBDTR \cite{tifs19vtreid}  &\scriptsize{TIFS19}& 27.82 & 67.34 & 81.34 & 28.42 & 32.46 & 77.42 & 89.62 & 42.46 \\
  HSME \cite{aaai19hsme} &\scriptsize{AAAI19}& 20.68 & 32.74 & 77.95  & 23.12 & - & -& -& - \\
  D$^2$RL \cite{cvpr19ivreid}  &\scriptsize{CVPR19}& 28.9 & 70.6  & 82.4 & 29.2  & - & -& -& - \\
  MAC \cite{tip20mace}      &\scriptsize{MM19}& 33.26  & 79.04 & 90.09   & 36.22       & 36.43   & 62.36  & 71.63 & 37.03\\
  MSR \cite{tip19msr} &\scriptsize{TIP19}& 37.35 & 83.40 & 93.34 & 38.11 &  39.64  & 89.29   & 97.66   &  50.88  \\
  AlignGAN \cite{iccv19ivreid}&\scriptsize{ICCV19} & 42.4 & 85.0 & 93.7 & 40.7 & 45.9 & 87.6 & 94.4 & 54.3 \\
  HPILN  \cite{arxiv19hpiln} &\scriptsize{arXiv19}& 41.36 & 84.78  & 94.31  & 42.95  &  45.77 & 91.82 & \textbf{98.46} & 56.52 \\
  LZM \cite{arxiv19efficient} &\scriptsize{arXiv19}& 45.00 & 89.06 & -&45.94 & 49.66    & 92.47  & - & 59.81 \\
  AGW \cite{arxiv20reidsurvey} &\scriptsize{arXiv20} & 47.50 & 84.39 & 92.14 & 47.65 & 54.17 & 91.14 & 95.98 & 62.97 \\
  Xmodal \cite{aaai20xmodal}&\scriptsize{AAAI20}& 49.92 & 89.79 & \textbf{95.96} &50.73 & -    & -  & - & - \\\hline
  DDAG (Ours)      & \centering{-} &  \textbf{54.75}   & \textbf{90.39} & \textbf{95.81}  & \textbf{53.02}  & \textbf{61.02}  & \textbf{94.06} & 98.41 & \textbf{67.98}\\\hline
 \end{tabular}
\end{table*}

\begin{table}\scriptsize
\centering
\caption{\label{tab:regdb}Comparison with the state-of-the-art methods on RegDB dataset on visible-infrared and infrared-visible settings. Rank at $r$ accuracy (\%) and mAP (\%) are reported.}
\begin{tabular}{l|P{1.1cm}P{1.1cm}P{1.1cm}|P{1.2cm}||P{1.1cm}P{1.1cm}P{1.1cm}|P{1.2cm}}
  \hline
    Setting&\multicolumn{4}{c||}{\emph{Visible to Infrared}} &\multicolumn{4}{c}{\emph{Infrared to Visible}} \\ \hline
  Method    & $r=1$   & $r=10 $   & $r=20$   & mAP  & $r=1$   & $r=10 $   & $r=20$   & mAP  \\\hline
  HCML \cite{aaai18vtreid}   & 24.44   &47.53  &  56.78 &20.08     & 21.70   &45.02  &  55.58 & 22.24   \\
  Zero-Pad  \cite{iccv17cross}   &17.75 &34.21   &44.35     & 18.90    & 16.63 & 34.68   &44.25     & 17.82    \\
  BDTR \cite{tifs19vtreid}      & 33.56  & 58.61   & 67.43     & 32.76      & 32.92  & 58.46   & 68.43    & 31.96   \\
  eBDTR \cite{tifs19vtreid}         & 34.62 & 58.96 & 68.72 & 33.46  & 34.21 & 58.74 & 68.64 & 32.49 \\
  HSME \cite{aaai19hsme} & 50.85 & 73.36 & 81.66  & 47.00  & 50.15 & 72.40 & 81.07  & 46.16 \\
  D$^2$RL \cite{cvpr19ivreid} & 43.4  & 66.1 & 76.3 & 44.1 & - & - & - & -\\
  MAC \cite{tip20mace} & 36.43   & 62.36  & 71.63 & 37.03 & 36.20  & 61.68   & 70.99     & 36.63 \\
  MSR \cite{tip19msr} & 48.43 & 70.32 & 79.95 & 48.67 & - & - & - & -\\
  EDFL \cite{arxiv19haijun} & 52.58   & 72.10  & 81.47 & 52.98  & 51.89  & 72.09  & 81.04 & 52.13  \\
  AlignGAN \cite{iccv19ivreid} & 57.9 & - & - &  53.6 & 56.3 & - & - &  53.4 \\
  Xmodal \cite{aaai20xmodal} & 62.21 & 83.13 & 91.72 & 60.18 & - & -& -& - \\\hline
  DDAG (Ours) & \textbf{69.34}    & \textbf{86.19} & \textbf{91.49} & \textbf{63.46} & \textbf{68.06}  & \textbf{85.15}  & \textbf{90.31} & \textbf{61.80}\\\hline
 \end{tabular}
\end{table}

\noindent\textbf{Parameter Analysis}
We evaluate the effect of different body parts $p$ and different numbers of graph attention heads $L$ on the large-scale SYSU-MM01 dataset, under the challenging \textit{all-search} mode. The results are shown in Fig. \ref{fig:param}.

(1) As shown in the left figure, a larger $p$ captures more fine-grained part features and improves the performance. However, when $p$ is too large, the performance drops since small body parts  cannot contain sufficient information for part attention learning. (2) As demonstrated in the right figure, a large $L$ provides more reliable relationship mining, and thus consistently improves the performance. However, it also greatly increases the difficulty of optimization, which results in a slightly decreased performance with a too large $L$. Thus, we select $p=3$ and $L=4$ in all our experiments.

\subsection{Comparison with State-of-the-Art Methods}\label{sec:compare}
This subsection presents a comparison with the current state-of-the-arts on two different datasets. The comparison includes eBDTR \cite{tifs19vtreid}, D$^2$RL \cite{cvpr19ivreid}, MAC \cite{tip20mace}, MSR \cite{tip19msr}, AlignGAN \cite{iccv19ivreid} and Xmodal \cite{aaai20xmodal}. Note that AlignGAN \cite{iccv19ivreid} represents the state-of-the-art by aligning the features in both the feature level and pixel level with generated images. Xmodal generates an intermediate modality to bridge the gap. We also compare with several arXiv papers, including EDFL \cite{arxiv19haijun}, HPILN \cite{arxiv19hpiln}, LZM \cite{arxiv19efficient} and AGW \cite{arxiv20reidsurvey}.
The results on two public datasets are shown in Tables \ref{tab:sysu} and \ref{tab:regdb}.

The following observations can be made: 1) Methods with two-stream networks (\textit{EDFL} \cite{arxiv19haijun}, \textit{MSR} \cite{tip19msr}, \textit{LZM }\cite{arxiv19efficient} and our proposed \textit{DDAG}) generally perform better than the one-stream network methods (\textit{cmGAN} \cite{ijcai18cross}, \textit{D$^2$RL} \cite{cvpr19ivreid} and \textit{Zero-Pad} \cite{iccv17cross}).  We
conjecture that the main reason is that two-stream networks can
simultaneously learn modality-specific and modality-sharable features,
which are more suitable for VI-ReID. 2) Our proposed DDAG significantly outperforms the current state-of-the-art AlignGAN \cite{iccv19ivreid} by a large margin on both datasets. Note that AlignGAN generates cross-modality image pairs to reduce the modality gap in both feature level and pixel level. In comparison, we do not require the time-consuming and resource-demanding image generation \cite{iccv19ivreid,cvpr19ivreid}, and our training process is quite efficient without the adversarial training \cite{ijcai18cross}, or the additional modality generation \cite{aaai20xmodal}.

Another experiment on the RegDB dataset (Table \ref{tab:regdb}) shows that DDAG is robust to different query settings. We achieve much better performance under both \textit{visible-to-infrared} and \textit{infrared-to-visible} query settings, suggesting that DDAG can learn better modality-sharable features by utilizing the part relationship within each modality and graph-structured  relations across two modalities.

\section{Conclusion}
We present a dynamic dual-attentive aggregation learning (DDAG) framework for VI-ReID. DDAG is innovative in two aspects: its IWPA component utilizes the part relationship within each modality to enhance the feature representation by simultaneously considering the part differences and relations; the CGSA module incorporates the neighborhood information across the two modalities to reduce the modality gap. We further design a dynamic dual aggregation learning strategy to seamlessly aggregate the two components. DDAG outperforms the state-of-the-art models on various settings, usually by a large margin. We believe the findings can also be applied in general single-modality person re-identification by mining the relation across multiple body parts, contextual images.

\bibliographystyle{splncs04}
\bibliography{ddag}
\end{document}